\documentclass[letterpaper, 10 pt, conference]{ieeeconf}  

\IEEEoverridecommandlockouts                              

\usepackage{cite}
\usepackage{fancyhdr, graphics, graphicx} 
\usepackage{epsfig} 
\usepackage{mathptmx} 
\usepackage{times} 
\usepackage{amsmath} 
\usepackage{amssymb}  
\usepackage{amsfonts}
\usepackage[ruled,vlined]{algorithm2e}
\usepackage{algorithmic}
\usepackage{bm}
\usepackage{booktabs}
\usepackage{dcolumn,tipa}
\newcolumntype{d}[1]{D{.}{.}{#1}}
\usepackage{etoolbox}
\AtBeginEnvironment{tcolorbox}{\small}
\usepackage{fixltx2e}
\usepackage{gensymb}
\usepackage{listings}
\usepackage{makecell}
\usepackage{multirow}
\usepackage{pifont}
\usepackage[most]{tcolorbox}
\usepackage{textcomp}
\usepackage{todonotes}
\usepackage{xcolor}
\usepackage{cleveref}
\usepackage{caption}
\usepackage{subcaption}
\DeclareMathOperator*{\argmax}{argmax}

\usepackage{xspace}
\newcommand{\etal}{\textit{et~al}.\xspace}
\tcbuselibrary{breakable}
\usepackage[activate={true,nocompatibility},final,tracking=true,kerning=true,spacing=true,factor=1100,stretch=10,shrink=10]{microtype}
\microtypesetup{protrusion=false}
\newcommand*\colorcheck[1]{%
  \expandafter\newcommand\csname #1check\endcsname{\textcolor{#1}{\ding{51}}}%
}
\newcommand*\colorcross[1]{%
  \expandafter\newcommand\csname #1cross\endcsname{\textcolor{#1}{\ding{55}}}%
}
\colorcheck{green}
\colorcross{red}

\renewcommand{\baselinestretch}{0.965}

\title{\LARGE \bf
Towards Human Awareness in Robot Task Planning with\\Large Language Models
}

\author{Yuchen Liu$^{1, 2}$, Luigi Palmieri$^{1}$, Sebastian Koch$^{1}$, Ilche Georgievski$^{2}$ and Marco Aiello$^{2}$
\thanks{$^{1}$Corporate Sector Research and Advance Engineering, Robert Bosch GmbH, Robert-Bosch-Campus 1, 71272 Renningen, Germany
        {\tt\small \{yuchen.liu2, luigi.palmieri, sebastian.koch2\}@de.bosch.com}}%
\thanks{$^{2}$Service Computing Department, Institute of Architecture of Application Systems, University of Stuttgart, Universitätsstraße 38, 70569 Stuttgart, Germany
        {\tt\small \{yuchen.liu, ilche.georgievski, marco.aiello\}@iaas.uni-stuttgart.de}}%
}

\begin{document}
\bstctlcite{IEEEexample:BSTcontrol}
\include{pythonlisting}

\maketitle
\thispagestyle{empty}
\pagestyle{empty}

\begin{abstract}
The recent breakthroughs in the research on Large Language Models (LLMs) have triggered a transformation across several research domains. Notably, the integration of LLMs has greatly enhanced performance in robot Task And Motion Planning (TAMP). However, previous approaches often neglect the consideration of dynamic environments, i.e., the presence of dynamic objects such as humans. In this paper, we propose a novel approach to address this gap by incorporating human awareness into LLM-based robot task planning. To obtain an effective representation of the dynamic environment, our approach integrates humans' information into a hierarchical scene graph. To ensure the plan's executability, we leverage LLMs to ground the environmental topology and actionable knowledge into formal planning language. Most importantly, we use LLMs to predict future human activities and plan tasks for the robot considering the predictions. Our contribution facilitates the development of integrating human awareness into LLM-driven robot task planning, and paves the way for proactive robot decision-making in dynamic environments.
\end{abstract}

\section{INTRODUCTION}
\begin{figure}[t]
    \centering
    \includegraphics[width=0.5\textwidth]{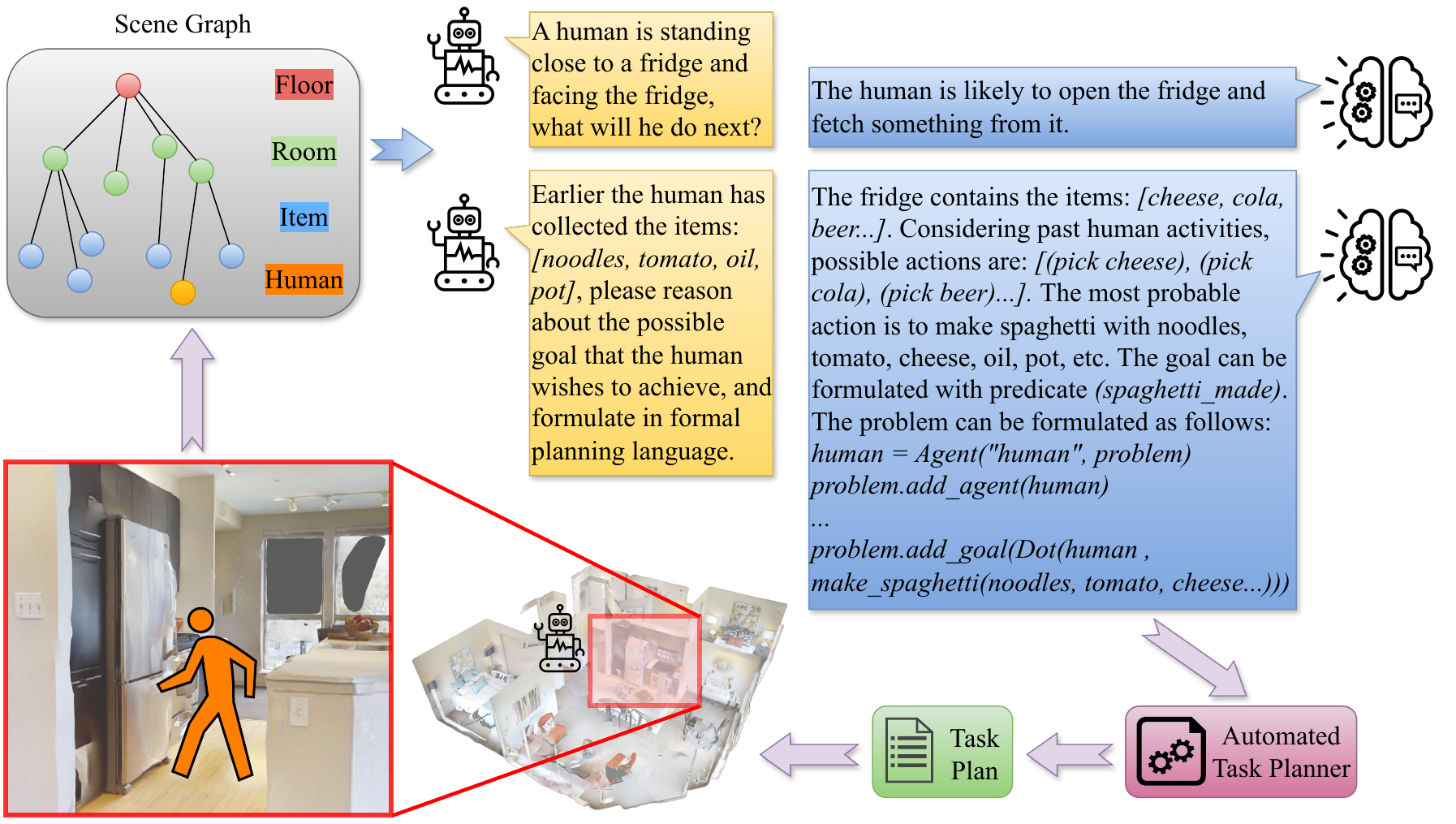}
    \caption{An example of the human-aware robot task planning upon the \textit{Allensville} environment \cite{Armeni2019hierachical}. A scene graph is pre-built from the environment with the floor, room, and item layers, where humans are modeled as nodes in the item layer. The edges refer to the semantic relationships. Using the scene graph as the environment representation, the human user queries a Large Language Model (LLM) to predict the next possible human activities and the goal states of the humans. An automated task planner generates a task plan considering humans as additional planning agents.\vspace{-1em}}
    \label{fig:fig1}
\end{figure}

In recent years, with robots playing more important roles in collaborations with humans in industrial, transportation, and household environments, planning robot tasks and motions while considering motions of cohabitating humans also becomes a crucial topic \cite{liu2023human, goel2023semantically}; e.g., a robot performing household tasks while avoiding disturbing the human in the vicinity, as shown in Fig.~\ref{fig:fig1}.
Meanwhile, a multitude of Large Language Models (LLMs) have emerged thanks to significant advancements in the research of Natural Language Processing (NLP), demonstrating the ability to generate human-like text, programming code, and service compositions with high proficiency \cite{openai2023gpt4, chowdhery2023palm, devlin2018bert, touvron2023llama, aiello2023service}.
Witnessing the capabilities of the LLMs, many researchers utilize them to tackle robot Task And Motion Planning (TAMP) problems \cite{huang2022language, song2023llm, liu2023llm+, ding2023task, ahn2022i, rana2023sayplan, liu2024delta}.
A popular strategy is to extract common-sense knowledge from the LLMs and use it as constraints of the classical automated task planning algorithms to improve the correctness and executability of the generated plans \cite{huang2022language, ding2023task, chen2023autotamp}. Other approaches also use LLMs to generate task specifications formulated in formal planning language, e.g., Planning Domain Definition Language (PDDL) \cite{mcdermott1998pddl}, such that the problem can be solved by the off-the-shelf automated task planners, as did in previous work~\cite{liu2023llm+, liu2024delta}.

For robots operating in large and complex environments, interpreting the underlying semantic information in the environment is an important factor leading to successful navigation \cite{goel2023semantically}. Recently, many studies have investigated the potential to improve the efficiency of navigation by encoding environment topology and semantic relations in high-level representations, e.g., scene graphs \cite{ravichandran2022hierarchical, agia2022taskography}. Particularly in high-level task planning, utilizing scene graphs can effectively enhance the planning performance by reducing the search space and the planning time \cite{rana2023sayplan, liu2024delta}.

Surprisingly, the research on incorporating human awareness into robot task planning has not seen meaningful progress recently. Early approaches mainly consider the human-aware task planning problems as scheduling problems, where humans have their own agenda, and the robot should plan its tasks by avoiding conflicts with humans' agenda \cite{alami2006toward, cirillo2010human, kockemann2016constraint}. However, most of these approaches rely on pre-provided humans' agendas. Despite the enormous advancement in LLMs and their wealth of common-sense knowledge, many previous LLM-based robot task planning approaches do not consider the presence of humans. Studies that leverage LLMs in human-aware task planning are yet rarely seen \cite{graule2023gg}.

Observing such a research gap, in this paper, we propose a novel approach to incorporate human awareness in robot task planning with LLMs. Our approach presents the following key contributions:

\emph{i)} We introduce a novel combination of scene graphs and LLMs. To enable human awareness, we encode humans with their semantic relationships to other static objects into scene graphs, and use LLMs to predict future human activities based on past observations of the relationships. 

\emph{ii)} Considering humans as additional planning agents with the predicted activities being their goals, our approach transforms the human-aware single-robot task planning problem into a multi-agent task planning problem. Utilizing LLMs to further ground the problem specifications into formal planning language, the approach can effectively solve the multi-agent problem jointly with the robot and humans while ensuring the plan's executability.

\emph{iii)} Our contribution facilitates the development of integrating human awareness into LLM-driven robot task planning, and paves the way for proactive robot decision-making in dynamic environments.

\begin{figure}[t]
    \centering
    \includegraphics[width=0.5\textwidth]{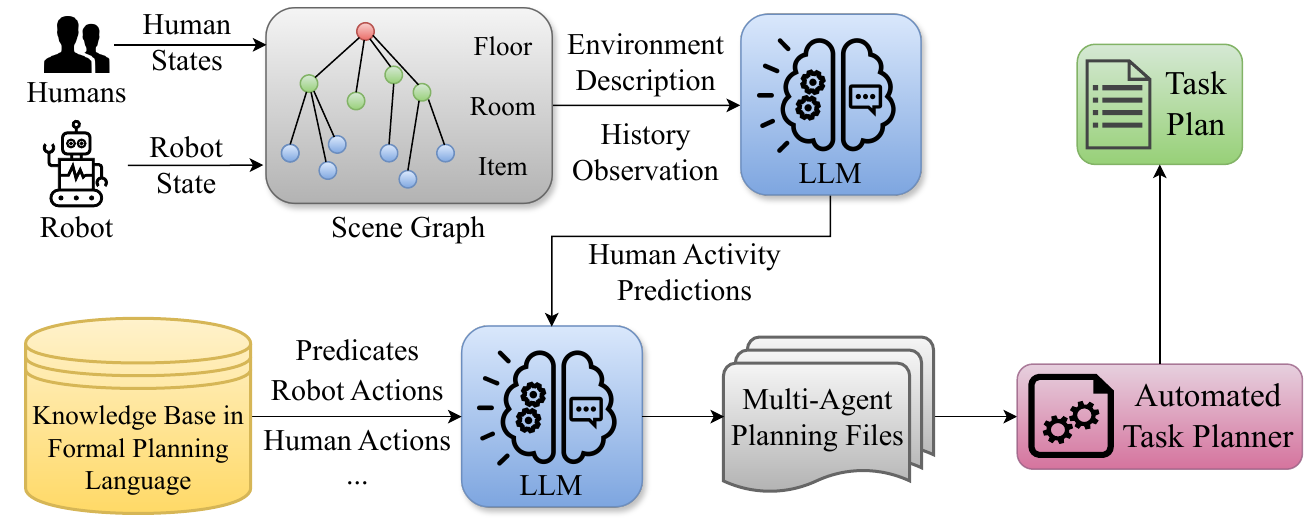}
    \caption{The architecture of the approach.\vspace{-1em}}
    \label{fig:sys}
\end{figure}

\section{RELATED WORK}\label{RW}

\subsection{3D Scene Graphs}
3D scene graphs are graph structures used to model large-scale physical environments. They offer an efficient hierarchical abstraction
for both spatial and semantic domains. 3D scene graphs were first introduced in \cite{Armeni2019hierachical} as a structure to connect buildings, rooms, objects, and cameras on multiple layers.  Following this introduction, Rosinol \etal \cite{rosinol2020rss, Rosinol_2021_SAGE} and Hughes \etal \cite{hughes2022hydra} investigated the construction of 3D scene graphs from sensor data in dynamic environments with humans. Wald \etal \cite{Wald2020D3SSG} introduce semantic 3D scene graphs that represent object relationships as edges between nodes representing the object instances in the scene.
Given their success, 3D scene graphs have started to be integrated into robotic systems for applications such as navigation \cite{seymour2022graphmapper,conceptgraphs,chang2023contextaware} or task and motion planning \cite{agia2022taskography,rana2023sayplan,liu2024delta}. 

\subsection{LLM-based Robot Task and Motion Planning}
In the past years, the development of incorporating LLMs into robot TAMP methodologies has achieved significant progress. A well-known approach involves leveraging LLMs to capture and employ their embedded rich semantic and common-sense knowledge to proficiently interpret the description of the environment written in Natural Language (NL) \cite{ding2023task, song2023llm}.
However, many researchers have already proven that the high-level task plans directly generated by LLMs are often incorrect and not executable since the actionable knowledge is not grounded specifically in the environment \cite{pallagani2024prospects}. Thus, various approaches are proposed for grounding LLMs' output into affordable actions \cite{ahn2022i, huang2022language, chen2023autotamp}, or into formal planning language such as PDDL for obtaining high-quality plans \cite{liu2023llm+, silver2022pddl} or even efficiently tackling long-term tasks \cite{liu2024delta}.

Understanding the semantic relations embedded in large environments is a crucial factor for the robots navigating in such scenarios. Scene graphs work as efficient environment representations thanks to the hierarchical structure and the compactly encoded semantic relationships \cite{chen2023open}. Utilizing LLMs to reason over scene graphs allows the comprehension of large environments, and enables a significant performance boost in robot task planning approaches in terms of the planning success rate \cite{rana2023sayplan, liu2024delta}.

\subsection{Human-aware Robot Task Planning}
A vast amount of studies in human-aware robot task planning have been introduced in the past decades \cite{alami2006toward, cirillo2010human, kockemann2016constraint, surma2021multiple, liu2023human, graule2023gg}. Classical approaches treat it as a scheduling problem with pre-provided or predicted human schedule, and plan the tasks for the robots while trying to avoid disturbing the humans \cite{cirillo2010human, kockemann2016constraint}. However, they only consider the presence of one human in the environment. To enable crowd awareness, several recent approaches consider the crowd as a whole entity and model its own long-term behaviors. Palmieri \etal \cite{palmieri2017kinodynamic}, Surma \etal \cite{surma2021multiple} and Liu \etal \cite{liu2023human} use Map of Dynamics (MoDs) to model the human motion patterns from past observations, and then plan tasks or motions for the robot considering the MoDs to achieve lower cost of traversing through human flow.

Nevertheless, to the best of our knowledge, despite the power of the LLMs, only one work contributes to LLM-driven human-aware robot task planning. Graule \etal \cite{graule2023gg} used a pre-trained LLM to predict the next probable human activity in terms of which object in the environment the human is likely to interact with, and plan the next actions for the robot to avoid disturbance to the human. However, they refer to the disturbance to the same-room occupation of the robot and the human, which is less realistic. Moreover, the applied task planning strategies are solely greedy policies. The vision of long-term planning is missing. Furthermore, they also consider one single human.
To summarize, integrating LLMs into human-aware robot task planning with multiple cohabitating humans still remains an open research topic.

\begin{figure}[t]
    \centering
    \includegraphics[width=0.3\textwidth]{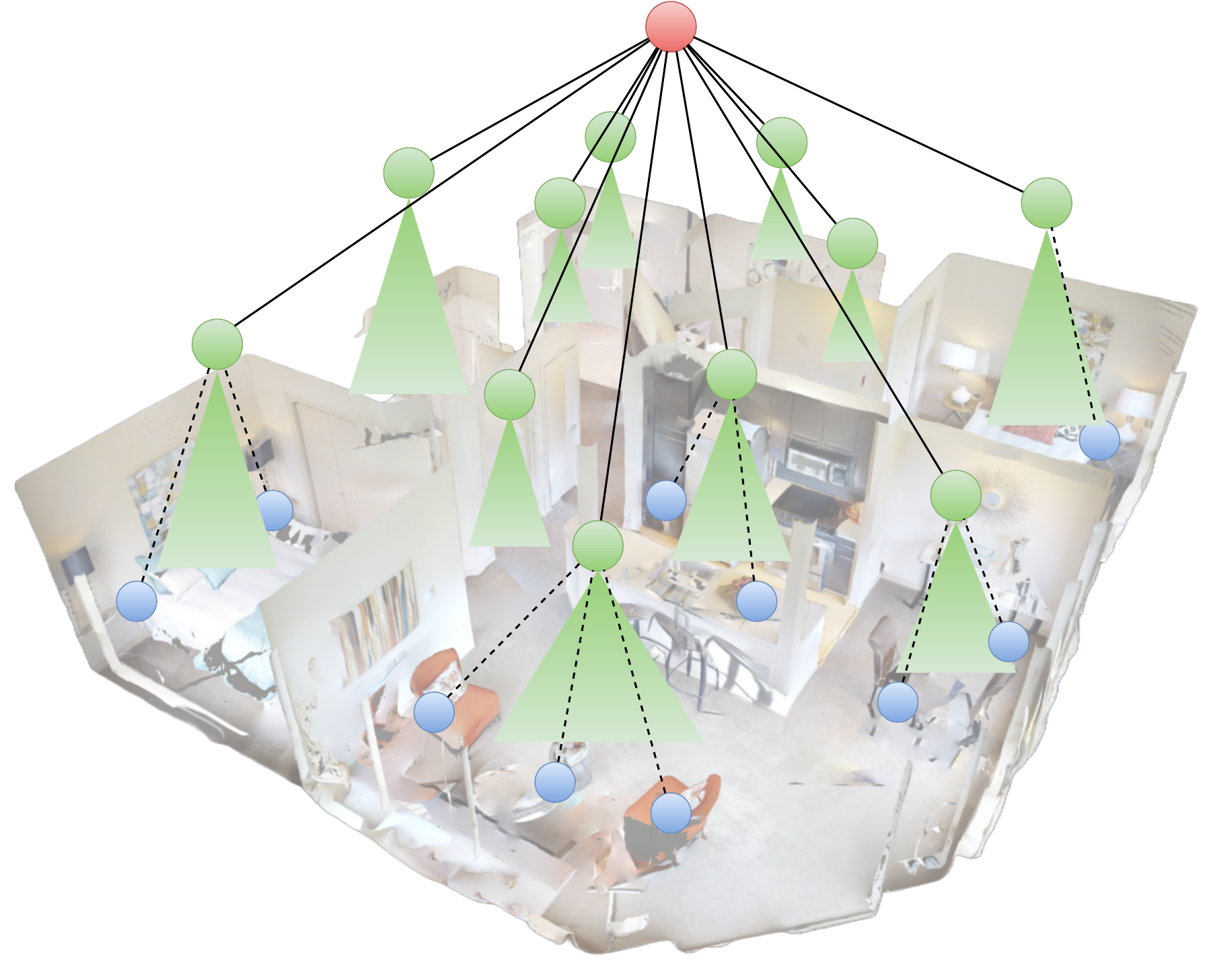}
    \caption{Illustration of the \textit{Allensville} scene and the corresponding 3D scene graph \cite{Armeni2019hierachical}. The scene graph has three layers: floor (red), room (green), and item (blue) layers. The humans and part of the item nodes are not visualized.\vspace{-1em}}
    \label{fig:sg}
\end{figure}

\section{APPROACH}\label{algo}
\subsection{Problem Statement}
We focus on solving human-aware robot task planning problems with LLMs. We consider a mobile robot that moves in household environments and performs different tasks. In the same space there are also multiple humans moving around. The robot should complete its own task while avoiding disturbing the humans. Given a 3D scene graph as the environment representation, and the knowledge base of the robot's and humans' actions, the LLM will predict future human activities and generate the task specifications,
such that an automated task planner can solve the problem.

In order to distinguish the \textit{object} keyword in formal planning language (e.g., PDDL) from objects in scene graphs, in the following, we refer to the scene graphs objects as \textit{items}. Additionally, we define \textit{agent} as an object type in the domain specification, while \textit{robot} and \textit{humans} as instances of \textit{agent} in the problem instances. Furthermore, we assume that the humans have rational actions, we also assume the full observability of all nodes in the scene graphs.

\subsection{Architecture}
Fig.~\ref{fig:sys} illustrates the system architecture. The input consists of the states of the robot and the humans, i.e., the position, the currently performing action, etc., a pre-built scene graph as the environment representation, and a knowledge base containing the domain knowledge, such as the object types and the affordable actions of the robot and the humans with preconditions and effects. Eventually, the approach returns a high-level task plan consisting of multiple symbolic actions. A simple example of a task plan looks like: \textit{(goto room A), (pick item X), (goto room B), (drop item X)}.

The scene graph has a hierarchical structure with three layers of nodes: floor, room, and item layers. Fig.~\ref{fig:sg} shows an example of the scene graph's layout with the static items. The room nodes are annotated with their neighboring rooms, and the item nodes contain several attributes, e.g., accessibility, states, and affordable actions.
The robot and humans are integrated into the scene graph and modeled as nodes in the item layer, as illustrated in Fig.~\ref{fig:fig1}. An important prerequisite for predicting future human activities is to predict the semantic relationships between the human node and the item nodes in the vicinity of the human. This can be done by utilizing existing methods such as \cite{Wald2020D3SSG} that predict such relationships and encode them as edges in scene graphs.

As the scene graph is formulated in programming code, after obtaining the extended scene graph with humans, it can be easily integrated into the prompt for the LLM for predicting future human activities with probabilities. For more accurate prediction, we take the history of the observations into account, i.e., the historical states of humans as well as the items that humans have interacted with in the past.

The core idea of our approach is to transform the human-aware single-robot task planning problem into a multi-agent task planning problem in such a way that humans are considered as additional planning agents, with the predicted activities as their goals, as did previously in many studies that consider human motion predictions as planning problems \cite{rudenko2020human}. Using the available predicates and the action knowledge from the knowledge base, the LLM can transform the predicted human activities into the goal states of humans and formulate them in formal planning language. As a result, we obtain the necessary multi-agent planning specifications. Querying an off-the-shelf automated task planner, the multi-agent task planning problem can be solved by generating a symbolic task plan.

\setlength{\textfloatsep}{4pt}
\begin{algorithm}[t]
\small
\DontPrintSemicolon
\LinesNumbered
\SetAlgoLined
\KwData{$domain$, $problem$ $\Pi$, $K$, $SG$, $LLM$}
\KwResult{$\pi$}
$object\_types, predicates, actions \leftarrow LLM(K.get\_knowledge(domain))$\; \label{alg:know_begin}
Add $object\_types$, $predicates$, and $actions$ to $domain$\; \label{alg:know_end}
$problem.add\_agent(SG.get\_robot\_node())$\;
\For{$n_{i} \in SG$}{
    $problem.set\_init\_state(n_{i})$\; \label{alg:s_item}
}
\For{$n_{h} \in SG_t$, $t \in [1, t_{n}], n \in \mathbb{Z}^+$}{ \label{alg:s_human_begin}
    $problem.add\_agent(n_{h})$\; 
    $E_{h}^{t}, I_{h}^{t} \leftarrow SG_t.get\_edges\_and\_neighbors(n_{h})$\;
    $\bm{H}_{h}^{1:t_{n}} \leftarrow \{E_{h}^{1:t_{n}} \cup I_{h}^{1:t_{n}}\}$\; \label{alg:s_human_end}
    $\{g_{h, m}^{t_{n+1}} \sim p_m \mid \sum_{m=1}^M p_m(g_{h, m}^{t_{n+1}} \mid \bm{H}_{h}^{1:t_{n}}) = 1, M \in \mathbb{Z}^+\} \leftarrow LLM(n_{h}, \bm{H}_{h}^{1:t_{n}})$\; \label{alg:llm_pred}
    \If{no predicates or actions correspond to $g_{h, 1:M}^{t_{n+1}}$}{ \label{alg:new_act_begin}
        $predicates'_{h}, actions'_{h} \leftarrow LLM(g_{h, 1:M}^{t_{n+1}})$\;
        Add $predicates'_{h}$ and $actions'_{h}$ to $domain$\;
    } \label{alg:new_act_end}
    $problem.add\_goal(n_{h}, \argmax_m  p_m(g_{h, m}^{t_{n+1}}))$\; \label{alg:goal}
}
$\pi \leftarrow \Pi(domain, problem)$\; \label{alg:plan}
\caption{Transforming Human Awareness into Multi-Agent Task Planning Problem with LLM}
\label{alg:llm_plan}
\end{algorithm}

\subsection{Transforming Human Awareness into Multi-Agent Task Planning Problem}
While automated planning techniques deliver optimal solutions, tackling task planning problems with dynamic objects such as humans still remains a challenging topic. Therefore, we propose a method to convert dynamic objects into planning agents. Knowing the action knowledge, the initial states, and the desired goal states of these agents, it is achievable to avoid conflicts between the robot and the other agents, i.e., humans, when planning jointly, thus enabling human awareness.

Algorithm \ref{alg:llm_plan} is about how the single-robot task planning problem is transformed into a multi-agent task planning problem. It takes the specifications of \textit{domain} and \textit{problem}, the automated task planner \textit{$\Pi$}, the knowledge base \textit{K}, the scene graph \textit{SG}, and the \textit{LLM} as inputs. Firstly, the \textit{LLM} extracts the domain knowledge about object types and actions from the knowledge base \textit{K}, formulates them in formal planning language, and adds to the domain specification (L. \ref{alg:know_begin}-\ref{alg:know_end}). Subsequently, the algorithm loops through the item nodes $n_i$ and the human nodes $n_h$ from the scene graph \textit{SG} considering all time steps $t \in [1, t_{n}]$. In the loop, the algorithm adds the initial states of each item (L. \ref{alg:s_item}), as well as the historical observations of the problem (L. \ref{alg:s_human_begin}-\ref{alg:s_human_end}), including the item nodes $I_{h}^{t}$ which the humans have interacted in the past, and the semantic relationships (i.e., edges $E_{h}^{t}$) between humans and the items. Based on past observations, the \textit{LLM} can predict $M$ numbers of future human activities in terms of the desired goal states $g_{h}^{t_{n+1}}$ for the humans with corresponding probabilities $p(g_{h}^{t_{n+1}} \mid \bm{H}_{h}^{1:t_{n}})$ (L. \ref{alg:llm_pred}). If there exist no predicates or actions in the domain that correspond to the predicted goal states, we can again leverage the rich common-sense knowledge from the \textit{LLM} to generate new ones and append them to the domain (L. \ref{alg:new_act_begin}-\ref{alg:new_act_end}), as we did in our previous work \cite{liu2024delta}.
Finally, we assign the goal with the highest probability to the corresponding human (L. \ref{alg:goal}). As a result, the task plan $\pi$ can be obtained by solving the problem with the planner $\Pi$ (L. \ref{alg:plan}).

\section{CONCLUSION}\label{conclusion}
We propose a novel approach for enabling human awareness in LLM-based robot task planning, where we integrate humans as nodes in a hierarchical scene graph for a unified environment representation, and use LLMs to predict future human activities based on the semantic relationships between humans and other static objects from the scene graph. A key contribution is the transformation of the single-robot task planning problem into a multi-agent problem, where humans are considered as additional planning agents with predicted activities as their goal states. Human awareness can be achieved by planning tasks for the robot jointly with humans to avoid disturbances.

We leave for future work to experiment with the proposed architecture and algorithm. In particular, we plan to implement and evaluate it against other competitive baselines with various tasks and scenarios, and also prepare experiments in photorealistic simulations and in real-world applications.

\section*{ACKNOWLEDGMENTS}
This work was partly supported by the EU Horizon 2020 research and innovation program under grant agreement No. 101017274 (DARKO).

\footnotesize
\bibliographystyle{IEEEtran}
\bibliography{main}

\def\authornoop#1{}
\begin{thebibliography}{10}
\providecommand{\url}[1]{#1}
\csname url@samestyle\endcsname
\providecommand{\newblock}{\relax}
\providecommand{\bibinfo}[2]{#2}
\providecommand{\BIBentrySTDinterwordspacing}{\spaceskip=0pt\relax}
\providecommand{\BIBentryALTinterwordstretchfactor}{4}
\providecommand{\BIBentryALTinterwordspacing}{\spaceskip=\fontdimen2\font plus
\BIBentryALTinterwordstretchfactor\fontdimen3\font minus \fontdimen4\font\relax}
\providecommand{\BIBforeignlanguage}[2]{{%
\expandafter\ifx\csname l@#1\endcsname\relax
\typeout{** WARNING: IEEEtran.bst: No hyphenation pattern has been}%
\typeout{** loaded for the language `#1'. Using the pattern for}%
\typeout{** the default language instead.}%
\else
\language=\csname l@#1\endcsname
\fi
#2}}
\providecommand{\BIBdecl}{\relax}
\BIBdecl

\bibitem{liu2023human}
Y.~Liu, L.~Palmieri, I.~Georgievski, and M.~Aiello, ``Human-flow-aware long-term mobile robot task planning based on hierarchical reinforcement learning,'' \emph{IEEE Robotics and Automation Letters}, 2023.

\bibitem{goel2023semantically}
Y.~Goel, N.~Vaskevicius, L.~Palmieri, N.~Chebrolu, K.~O. Arras, and C.~Stachniss, ``Semantically informed mpc for context-aware robot exploration,'' in \emph{2023 IEEE/RSJ International Conference on Intelligent Robots and Systems (IROS)}.\hskip 1em plus 0.5em minus 0.4em\relax IEEE, 2023, pp. 11\,218--11\,225.

\bibitem{openai2023gpt4}
J.~Achiam \emph{et~al.}, ``Gpt-4 technical report,'' \emph{arXiv preprint arXiv:2303.08774}, 2023.

\bibitem{chowdhery2023palm}
A.~Chowdhery \emph{et~al.}, ``Palm: Scaling language modeling with pathways,'' \emph{Journal of Machine Learning Research}, vol.~24, no. 240, pp. 1--113, 2023.

\bibitem{devlin2018bert}
J.~Devlin, M.-W. Chang, K.~Lee, and K.~Toutanova, ``Bert: Pre-training of deep bidirectional transformers for language understanding,'' \emph{arXiv preprint arXiv:1810.04805}, 2018.

\bibitem{touvron2023llama}
H.~Touvron \emph{et~al.}, ``Llama: Open and efficient foundation language models,'' \emph{arXiv preprint arXiv:2302.13971}, 2023.

\bibitem{aiello2023service}
M.~Aiello and I.~Georgievski, ``Service composition in the chatgpt era,'' \emph{Service Oriented Computing and Applications}, pp. 1--6, 2023.

\bibitem{huang2022language}
W.~Huang, P.~Abbeel, D.~Pathak, and I.~Mordatch, ``Language models as zero-shot planners: Extracting actionable knowledge for embodied agents,'' in \emph{International Conference on Machine Learning}.\hskip 1em plus 0.5em minus 0.4em\relax PMLR, 2022, pp. 9118--9147.

\bibitem{song2023llm}
C.~H. Song, J.~Wu, C.~Washington, B.~M. Sadler, W.-L. Chao, and Y.~Su, ``Llm-planner: Few-shot grounded planning for embodied agents with large language models,'' in \emph{Proceedings of the IEEE/CVF International Conference on Computer Vision}, 2023, pp. 2998--3009.

\bibitem{liu2023llm+}
B.~Liu \emph{et~al.}, ``Llm+ p: Empowering large language models with optimal planning proficiency,'' \emph{arXiv preprint arXiv:2304.11477}, 2023.

\bibitem{ding2023task}
Y.~Ding, X.~Zhang, C.~Paxton, and S.~Zhang, ``Task and motion planning with large language models for object rearrangement,'' \emph{arXiv preprint arXiv:2303.06247}, 2023.

\bibitem{ahn2022i}
M.~Ahn \emph{et~al.}, ``Do as i can, not as i say: Grounding language in robotic affordances,'' \emph{arXiv preprint arXiv:2204.01691}, 2022.

\bibitem{rana2023sayplan}
K.~Rana, J.~Haviland, S.~Garg, J.~Abou-Chakra, I.~Reid, and N.~Suenderhauf, ``Sayplan: Grounding large language models using 3d scene graphs for scalable robot task planning,'' in \emph{Conference on Robot Learning}.\hskip 1em plus 0.5em minus 0.4em\relax PMLR, 2023, pp. 23--72.

\bibitem{liu2024delta}
Y.~Liu, L.~Palmieri, S.~Koch, I.~Georgievski, and M.~Aiello, ``Delta: Decomposed efficient long-term robot task planning using large language models,'' \emph{arXiv preprint arXiv:2404.03275}, 2024.

\bibitem{chen2023autotamp}
Y.~Chen, J.~Arkin, Y.~Zhang, N.~Roy, and C.~Fan, ``Autotamp: Autoregressive task and motion planning with llms as translators and checkers,'' \emph{arXiv preprint arXiv:2306.06531}, 2023.

\bibitem{mcdermott1998pddl}
D.~McDermott \emph{et~al.}, ``Pddl-the planning domain definition language,'' 1998.

\bibitem{ravichandran2022hierarchical}
Z.~Ravichandran, L.~Peng, N.~Hughes, J.~D. Griffith, and L.~Carlone, ``Hierarchical representations and explicit memory: Learning effective navigation policies on 3d scene graphs using graph neural networks,'' in \emph{2022 International Conference on Robotics and Automation (ICRA)}.\hskip 1em plus 0.5em minus 0.4em\relax IEEE, 2022, pp. 9272--9279.

\bibitem{agia2022taskography}
C.~Agia \emph{et~al.}, ``Taskography: Evaluating robot task planning over large 3d scene graphs,'' in \emph{Conference on Robot Learning}.\hskip 1em plus 0.5em minus 0.4em\relax PMLR, 2022.

\bibitem{Armeni2019hierachical}
I.~Armeni \emph{et~al.}, ``3d scene graph: A structure for unified semantics, 3d space, and camera,'' in \emph{Proceedings of the IEEE/CVF International Conference on Computer Vision (ICCV)}, 2019.

\bibitem{alami2006toward}
R.~Alami, A.~Clodic, V.~Montreuil, E.~A. Sisbot, and R.~Chatila, ``Toward human-aware robot task planning,'' in \emph{AAAI spring symposium: to boldly go where no human-robot team has gone before}, 2006.

\bibitem{cirillo2010human}
M.~Cirillo, L.~Karlsson, and A.~Saffiotti, ``Human-aware task planning: An application to mobile robots,'' \emph{ACM Transactions on Intelligent Systems and Technology (TIST)}, vol.~1, no.~2, pp. 1--26, 2010.

\bibitem{kockemann2016constraint}
U.~K{\"o}ckemann, ``Constraint-based methods for human-aware planning,'' Ph.D. dissertation, {\"O}rebro university, 2016.

\bibitem{graule2023gg}
M.~A. Graule and V.~Isler, ``Gg-llm: Geometrically grounding large language models for zero-shot human activity forecasting in human-aware task planning,'' \emph{arXiv preprint arXiv:2310.20034}, 2023.

\bibitem{rosinol2020rss}
A.~Rosinol, A.~Gupta, M.~Abate, J.~Shi, and L.~Carlone, ``3d dynamic scene graphs: Actionable spatial perception with places, objects, and humans,'' in \emph{Robotics: Science and Systems (RSS)}, 2020.

\bibitem{Rosinol_2021_SAGE}
A.~Rosinol \emph{et~al.}, ``Kimera: From slam to spatial perception with 3d dynamic scene graphs,'' \emph{The International Journal of Robotics Research}, vol.~40, no. 12-14, pp. 1510--1546, 2021.

\bibitem{hughes2022hydra}
N.~Hughes, Y.~Chang, and L.~Carlone, ``Hydra: A real-time spatial perception system for {3D} scene graph construction and optimization,'' 2022.

\bibitem{Wald2020D3SSG}
J.~Wald, H.~Dhamo, N.~Navab, and F.~Tombari, ``Learning 3d semantic scene graphs from 3d indoor reconstructions,'' in \emph{Proceedings of the IEEE/CVF Conference on Computer Vision and Pattern Recognition (CVPR)}, 2020.

\bibitem{seymour2022graphmapper}
Z.~Seymour, N.~C. Mithun, H.-P. Chiu, S.~Samarasekera, and R.~Kumar, ``Graphmapper: Efficient visual navigation by scene graph generation,'' in \emph{2022 26th International Conference on Pattern Recognition (ICPR)}.\hskip 1em plus 0.5em minus 0.4em\relax IEEE, 2022, pp. 4146--4153.

\bibitem{conceptgraphs}
Q.~Gu \emph{et~al.}, ``Conceptgraphs: Open-vocabulary 3d scene graphs for perception and planning,'' \emph{arXiv preprint arXiv:2309.16650}, 2023.

\bibitem{chang2023contextaware}
H.~Chang \emph{et~al.}, ``Context-aware entity grounding with open-vocabulary 3d scene graphs,'' in \emph{7th Annual Conference on Robot Learning}, 2023.

\bibitem{pallagani2024prospects}
V.~Pallagani \emph{et~al.}, ``On the prospects of incorporating large language models (llms) in automated planning and scheduling (aps),'' \emph{arXiv preprint arXiv:2401.02500}, 2024.

\bibitem{silver2022pddl}
T.~Silver, V.~Hariprasad, R.~S. Shuttleworth, N.~Kumar, T.~Lozano-P{\'e}rez, and L.~P. Kaelbling, ``Pddl planning with pretrained large language models,'' in \emph{NeurIPS 2022 Foundation Models for Decision Making Workshop}, 2022.

\bibitem{chen2023open}
B.~Chen \emph{et~al.}, ``Open-vocabulary queryable scene representations for real world planning,'' in \emph{2023 IEEE International Conference on Robotics and Automation (ICRA)}.\hskip 1em plus 0.5em minus 0.4em\relax IEEE, 2023, pp. 11\,509--11\,522.

\bibitem{surma2021multiple}
F.~Surma, T.~P. Kucner, and M.~Mansouri, ``Multiple robots avoid humans to get the jobs done: An approach to human-aware task allocation,'' in \emph{2021 European Conference on Mobile Robots (ECMR)}.\hskip 1em plus 0.5em minus 0.4em\relax IEEE, 2021, pp. 1--6.

\bibitem{palmieri2017kinodynamic}
L.~Palmieri, T.~P. Kucner, M.~Magnusson, A.~J. Lilienthal, and K.~O. Arras, ``Kinodynamic motion planning on gaussian mixture fields,'' in \emph{2017 IEEE International Conference on Robotics and Automation (ICRA)}.\hskip 1em plus 0.5em minus 0.4em\relax IEEE, 2017, pp. 6176--6181.

\bibitem{rudenko2020human}
A.~Rudenko, L.~Palmieri, M.~Herman, K.~M. Kitani, D.~M. Gavrila, and K.~O. Arras, ``Human motion trajectory prediction: A survey,'' \emph{The International Journal of Robotics Research}, vol.~39, no.~8, pp. 895--935, 2020.

\end{thebibliography}

\end{document}